\documentclass[10pt,twocolumn]{article}

\usepackage[accsupp]{axessibility}
\usepackage{wacv}
\usepackage{times}
\usepackage{epsfig}
\usepackage{graphicx}
\usepackage{amsmath}
\usepackage{amssymb}
\usepackage{booktabs}
\usepackage{multirow}
\usepackage{stfloats}
\usepackage{enumitem,amssymb}
\usepackage{amsfonts}
\usepackage{bbm}
\usepackage{array}
\usepackage{pdfpages}

\makeatletter
\newcommand{\thickhline}{%
    \noalign {\ifnum 0=`}\fi \hrule height 1pt
    \futurelet \reserved@a \@xhline
}

\def\wacvPaperID{***} 

\wacvfinalcopy 

\ifwacvfinal
\fi

\usepackage[pagebackref=true,breaklinks=true,colorlinks,bookmarks=false]{hyperref}

\ifwacvfinal
\usepackage[breaklinks=true,bookmarks=false]{hyperref}
\else
\usepackage[pagebackref=true,breaklinks=true,colorlinks,bookmarks=false]{hyperref}
\fi

\begin{document}

\title{What Makes for Effective Few-shot Point Cloud Classification?}

\author{
Chuangguan Ye\textsuperscript{1},
Hongyuan Zhu\textsuperscript{2}\thanks{This work was completed under the supervision of Dr. Hongyuan Zhu at A*STAR.}, 
Yongbin Liao\textsuperscript{1}, 
Yanggang Zhang\textsuperscript{1}, 
Tao Chen\textsuperscript{1}\thanks{Tao Chen is the corresponding author.}, 
Jiayuan Fan\textsuperscript{3}  \\
\textsuperscript{1}School of Information Science and Technology, Fudan University, China\\
\textsuperscript{2}Institute for Infocomm Research, A*STAR, Singapore \\
\textsuperscript{3}Academy for Engineering and Technology, Fudan University, China\\
{\tt\small\{cgye19,19210720121,ygzhang19,eetchen,jyfan\}@fudan.edu.cn, hongyuanzhu.cn@gmail.com}
}

\maketitle
\thispagestyle{empty}

\begin{abstract}
    Due to the emergence of powerful computing resources and large-scale annotated datasets, deep learning has seen wide applications in our daily life. However, most current methods require extensive data collection and retraining when dealing with novel classes never seen before. On the other hand, we humans can quickly recognize new classes by looking at a few samples, which motivates the recent popularity of few-shot learning (FSL) in machine learning communities. Most current FSL approaches work on 2D image domain, however, its implication in 3D perception is relatively under-explored. Not only needs to recognize the unseen examples as in 2D domain, 3D few-shot learning is more challenging with unordered structures, high intra-class variances, and subtle inter-class differences. Moreover, different architectures and learning algorithms make it difficult to study the effectiveness of existing 2D methods when migrating to the 3D domain.
    
    In this work, for the first time, we perform systematic and extensive studies of recent 2D FSL and 3D backbone networks for benchmarking few-shot point cloud classification, and we suggest a strong baseline and learning architectures for 3D FSL. Then, we propose a novel plug-and-play component called Cross-Instance Adaptation (CIA) module, to address the high intra-class variances and subtle inter-class differences issues, which can be easily inserted into current baselines with significant performance improvement. Extensive experiments on two newly introduced benchmark datasets, ModelNet40-FS and ShapeNet70-FS, demonstrate the superiority of our proposed network for 3D FSL. 
\end{abstract}
\vspace{-15pt}

\section{Introduction}
\label{sec:introduction}
    3D point cloud object classification is an important task for various computer vision and is widely  applied in many scenarios, like robotics~\cite{rusu2008towards}, indoor simultaneous localization and mapping (SLAM)~\cite{zhu2017target}, autonomous vehicles (AVs)~\cite{shi2020point,liang2018deep}, etc. unlike traditional point cloud recognition algorithms that extract handcrafted features~\cite{johnson1999using,zhong2009intrinsic,rusu2008aligning,tombari2010unique}, deep-learning based methods can learn more representative features from shape projections ~\cite{yu2018multi,qi2016volumetric,feng2018gvcnn} or raw points~\cite{qi2017pointnet,zhao2019pointweb,liu2019relation,liu2019densepoint} with deep networks, achieving better performance in kinds of point cloud processing tasks. 
    
    \begin{figure}[tp]
    \begin{center}
    \includegraphics[width=0.9\linewidth]{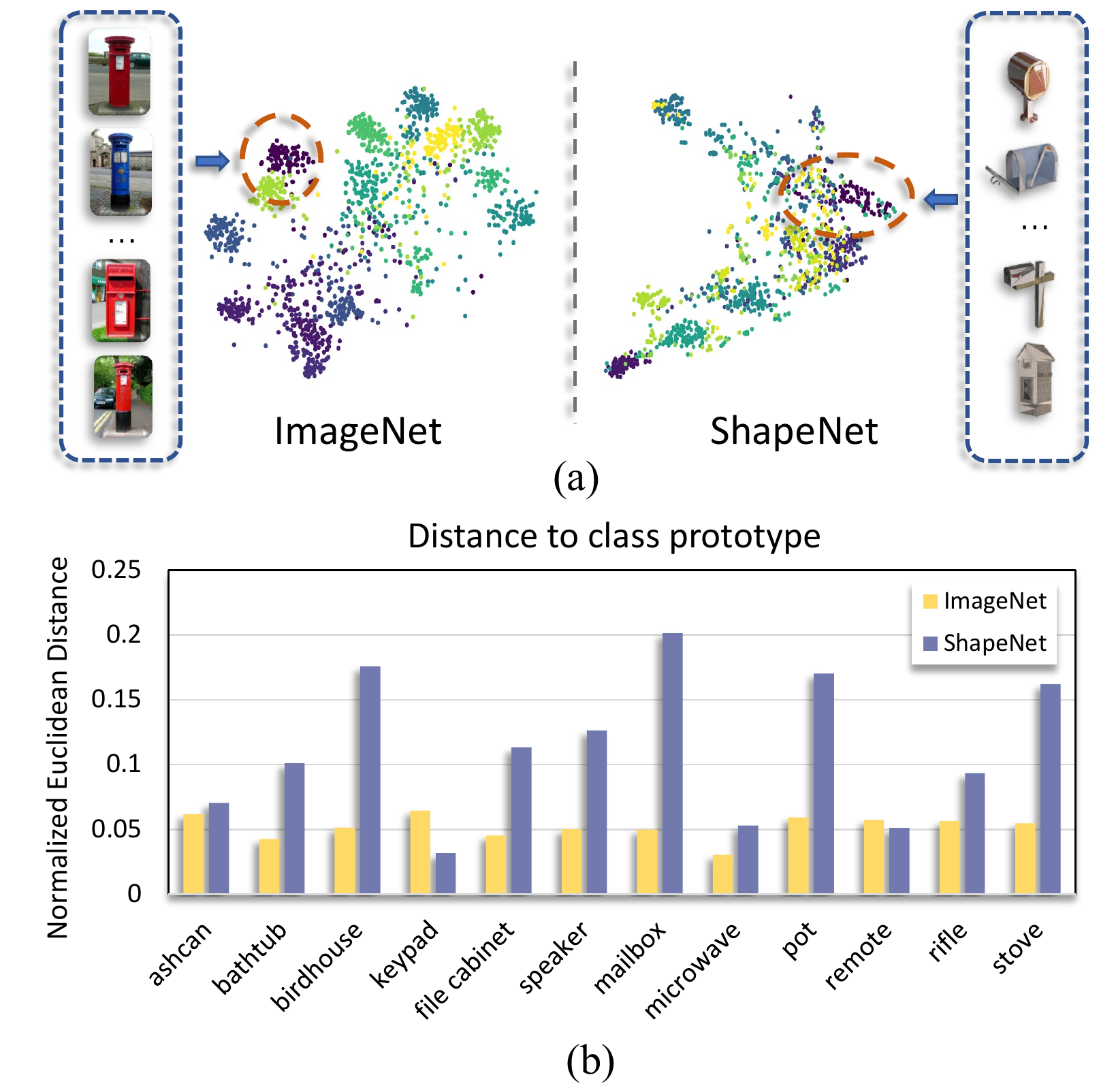}
    \end{center}
    \vspace{-10pt}
    \caption{The challenge of few-shot 3D point cloud classification. (a) is the T-SNE on ImageNet~\cite{Imagenet} and ShapeNet~\cite{ShapeNet} using pre-trained 2D image and 3D point cloud feature respectively with the \textbf{Mailbox} class highlighted. (b) The mean normalized Euclidean distance between each example and its class prototype in the visual and point cloud embedding space respectively. The embedding quality of 2D visual domain is much higher than 3D point cloud domains because image-based pre-trained models~\cite{TZSL}, such as ResNet, use deeper networks trained on millions of images, whereas point cloud-based models, such as PointNet, use shallower networks trained on only a few hundred point clouds with high intra-class variations and subtle inter-class differences.}
    \label{fig:challenge}
    \vspace{-15pt}
    \end{figure}

    However, there are two crucial problems of deep learning-based approaches for point cloud classification. Firstly, deep networks are usually data-driven and strongly rely on large amounts of labeled training data, while the process of annotating data is cumbersome and costly. Secondly, the networks tend to have poor generalization ability for novel classes never seen before.
    Data-augmentation based methods~\cite{salamon2017deep,cubuk2019autoaugment,ratner2017learning} and regularization techniques~\cite{hariharan2017low,yu2009deep}  can alleviate the data deficiency problem without adding new labeled data. But these methods may not get promising results on unseen classes or new tasks without sufficient labeled training data. In recent years, few-shot learning (FSL) algorithms~\cite{vinyals2016matching,PN} are introduced to rapidly generalize deep networks to new tasks with only one or few annotated examples, having great success in the 2D image domain. By constructing many auxiliary training meta-tasks, these 2D FSL algorithms use metric learning~\cite{PN,vinyals2016matching} or optimization-based approaches~\cite{MetaOpt} to learn transferable knowledge and propagate them to new tasks.
    
    Compared to the FSL in the 2D image domain, its study on 3D data is still relatively under-explored with the following challenges: 1) point cloud is a set of irregular, unstructured and unordered points defined in European space~\cite{bello2020deep}, to what extent existing point cloud architecture and learning algorithm can perform for 3D FSL is unknown; 2) most 2D FSL algorithms can learn more discriminative representations using deeper networks pre-trained on large-scale base class data~\cite{closerlook, TZSL}, e.g. tieredImageNet~\cite{ren2018meta} contains 608 classes and has a total number of 800,000 samples. By contrast, most 3D model datasets, e.g. ModelNet~\cite{ModelNet} and ShapeNet~\cite{ShapeNet}, have a much smaller number of labeled data (ModelNet~\cite{ModelNet} includes 40 classes with a total number of 12,311 samples). So the point-based models trained on low volume of data may generate poor-quality 3D feature clusters with high intra-class variations and subtle inter-class differences, as shown in Figure \ref{fig:challenge}. Therefore, how to address these issues requires further exploration. 
    
    In this paper, we study 3D few-shot learning (3D FSL) in a systematic manner for the first time. We perform extensive studies and discussions of various state-of-the-art point cloud architectures and few-shot learning algorithms in the context of 3D FSL, and propose a strong baseline for 3D FSL. Furthermore, to address the issues of high intra-class variance and subtle inter-class differences, we propose a plug-and-play Cross-Instance Adaption (CIA) module for 3D FSL. The CIA contains two modules called Self-Channel Interaction (SCI) module and Cross-Instance Fusion (CIF) module, which can be flexibly inserted into most current FSL algorithms with few changes and achieves significant performance improvement. Moreover, for objectively evaluating, we introduce new splits of ModelNet~\cite{ModelNet} and ShapeNetCore~\cite{ShapeNet} respectively, and construct two benchmark datasets, ModelNet40-FS, ShapeNet70-FS for 3D few-shot point cloud classification. Code and datasets are available at \url{https://github.com/cgye96/A\_Closer\_Look\_At\_3DFSL}.
    
    Our contributions can be summarized as:
    \begin{itemize}
    \vspace{-5pt}
        \item We are the first to perform a systematic study of 3D few-shot learning (FSL) in terms of network architectures and propose a strong baseline for 3D FSL. \vspace{-5pt}
        \item We propose a novel plug-and-play Cross Instance Adaption (CIA) module that can be flexibly inserted into most current FSL algorithms/backbones and achieves significant improvement for 3D point cloud few-shot object classification. \vspace{-5pt}
        \item The proposed method achieves state-of-the-art performance on the two newly built 3D FSL benchmark datasets ModelNet40-FS and ShapeNet70-FS under different settings.
    \end{itemize}
\vspace{-13pt}

\section{Problem Definition and Related Work}
\label{sec:problemdefinition}
    Let $(x, y)$ denote a point cloud instance $x$ and its label $y$, where $x = \{p_j|j=1,...,n\}$ is an unordered point set including $n$ points, and each point $p_j$ is represented by its 3D coordinates. In  standard $N$-way-$K$-shot FSL setting~\cite{closerlook}, the goal of FSL is to train a predictor for query examples with few labeled support examples, where the labeled support examples are denoted as support set $\mathcal{S} = \{(x_i,y_i)\}_{i=1}^{N_s= N \times K}$ containing $N$ classes with $K$ examples for each class, and the query examples are denoted as the query set $\mathcal{Q}  = \{(x_i,y_i)\}_{i=1}^{N_q=N \times Q}$ containing the same $N$ classes with $Q$ examples for each class. 
    
    For few-shot point cloud classification, meta-learning is the state-of-the-art paradigm with a set of meta-training episodes $\mathcal{T}=\{(\mathcal{S}_{i}, \mathcal{Q}_{i})\}_{i=1}^{I}$ by optimizing following objectives:
    \begin{equation}
	    \theta^*,\phi^* =\underset{\theta ,\phi}{\arg\min}\mathcal{L}\left(\mathcal{T};\theta ,\phi \right). 
	    \label{eq:optimization}
    \end{equation}
    where $\mathcal{T}$ are sampled from the training set and $\mathcal{L}$ denotes the cross-entropy loss function defined as:
     \vspace{-2pt}
    \begin{equation}
	    \mathcal{L}\left( \mathcal{T};\theta ,\phi \right) =\mathbb{E}_\mathcal{T}\left[ -\log p\left( \hat{y}=c|x \right) \right],
	    \label{eq:cross_entropy}
	     \vspace{-2pt}
    \end{equation}
    with the prediction  $p\left( \hat{y}=c|x \right)$ can be given by:
    \vspace{-2pt}
    \begin{equation}
        p\left( \hat{y}=c|x \right) =softmax\left( \mathcal{C}_{\theta}\left( \mathcal{F}_{\phi} \left( x \right) \right) \right), 
        \label{eq:prediction}
        \vspace{-2pt}
    \end{equation}
    where $x$ is the input point cloud instance, $\hat{y}$ is the predicted label. The $\mathcal{F}$ is the embedding network parameterized by $\phi$ and $\mathcal{C}$ is the classifier parameterized by $\theta$.
    
    Once meta-training is finished, the generalization of the predictor is evaluated on meta-testing episodes $\mathcal{V}=\{( \mathcal{S}_{j},\mathcal{Q}_{j})\}_{j=1}^{J}$, sampled from the testing set. Note that we denote the classes in $\mathcal{V}$ as novel classes, which are disjoint with the base classes in meta-training episodes $\mathcal{T}$.

    Therefore, there are two key challenges in Point Cloud FSL:
    1) how to represent the point cloud data properly for few-shot learning; 2) how to effectively transfer the knowledge gained in meta-training episodes to meta-testing episodes with a small number of labeled samples with high intra-class variance and subtle inter-class differences. In the following section, we'll introduce the recent efforts in 3D point cloud classification and 2D few-shot learning.

\subsection{Related Work}
\label{sec:relatedwork}
    {\bf 3D point cloud classification:} Unlike conventional handcrafted feature extraction approaches~\cite{tombari2010unique,johnson1999using}, deep-learning based methods can learn more complicated and representative features with deep networks. According to the way of feature extraction from structured grids or raw points, existing methods can be generally divided into projection-based and point-based. Projection-based networks first convert the irregular points into a structured representation, such as multi-view projection~\cite{su2015multi,yu2018multi}, voxel ~\cite{maturana2015voxnet,riegler2017octnet} or lattices~\cite{su2018splatnet,rao2019spherical}, and then apply the typical 2D or 3D convolution neural networks (CNN) to extract view-wise or structural features. However, these methods may suffer from explicit information loss or higher memory consumption ~\cite{guo2020deep}. 
    
    In contrast, directly applying deep networks on raw points is becoming a trend in recent years. PointNet ~\cite{qi2017pointnet} is the first deep network for an unstructured point cloud, which learns point-wise features with multilayer perceptron (MLP) and aggregates the global feature with a symmetric function implemented by a max-pooling layer. After that, lots of works such as PointNet++~\cite{qi2017pointnet++}, PointCNN~\cite{li2018pointcnn},  RSCNN~\cite{liu2019relation}, DensePoint~\cite{liu2019densepoint}, and DGCNN~\cite{wang2019dynamic} explore the local relations in a specific region with convolution-based networks or graph-based networks, which can capture meaningful geometric features and achieve the state-of-the-art performance. Nonetheless, these deep learning approaches are greedy for the availability of massive annotated data, and may have poor generalization on novel classes that are unseen during training. 

    Recently, there are some works that also consider 3D point cloud learning with a small number of training data. Sharma \emph{et al.}~\cite{SSLFSL} mainly studies feature representation learning using self-supervision which is orthogonal to our few-shot learning.  LSSB~\cite{lssb} try to learn a discriminative embedding space for 3D model multi-view images with the bias of point cloud shape.
    
    Different from these works, we are the first to study few-shot point cloud classification in a systematic manner by reviewing recent 3D point cloud learning networks, and suggest strong baselines for the problem. We also propose an effective plug-and-play module, Cross Instance Adaption module, to address the high intra-class variance and subtle inter-class differences issues of 3D FSL, which can be flexibly inserted into most current FSL algorithms.

    {\bf 2D few-shot learning:} With the characteristics of less annotated training data and good generalization on new tasks, few-shot learning is a promising direction for deep learning. In general, existing FSL algorithms are based on a meta-learning framework and can be roughly categorized into metric-based methods~\cite{vinyals2016matching,PN,sung2018learning,li2019revisiting, oreshkin2018tadam} and optimization-based methods~\cite{finn2017model,MetaOpt, ravi2016optimization, nichol2018first,rusu2018meta}. 

    Metric-based methods focus on learning an embedding space where similar sample pairs are closer, or designing a metric function to compare the feature similarity of samples. Matching Network~\cite{vinyals2016matching}adapts a bidirectional LSTM module to get full context embeddings and uses cosine distance to classify query samples. Prototypical Network~\cite{PN} on the other hand first averages the support-set features for each class as a class prototype and then takes squared Euclidean distance to measure the similarity with query samples, which demonstrates better performance than MatchingNet. Relation Net~\cite{sung2018learning} further proposes a learnable metric module to get relation scores between the support set and the query set. 

    Optimization-based methods, on the other hand, regard meta-learning as an optimization process. MAML~\cite{finn2017model} learns a model-agnostic initialization parameter that can produce great improvement with a few gradient steps on new tasks. MetaOptNet~\cite{MetaOpt} incorporates a differentiable quadratic programming solver to learn a feature-relevant linear SVM predictor which can offer better generalization for novel categories. 
    
    In this work, we perform a systematic study of different meta-learning algorithms under few-shot point cloud classification tasks, and suggest strong baselines and components for the problem.

\vspace{-5pt}
\section{Empirical Study}
\label{sec:empirical}
    In the following sections, we would like to perform an empirical study of recent state-of-the-art few-shot learning methods on point cloud data with different popular point-based network backbones on our newly proposed ModelNet40-FS and ShapeNet70-FS datasets. The experimental settings and implementation details can be found in Section~\ref{sec:experiments}.
    
\subsection{State-of-the-art 2D FSL on Point Cloud}
    We first analyze recent state-of-the-art 2D few-shot learning methods' performance, when they migrate to the few-shot point cloud classification task on our newly proposed benchmark datasets. Given its simplicity and efficiencies, we adopt PointNet~\cite{qi2017pointnet} as the backbone for feature embeddings.  Specifically, we divide the state-of-the-arts into the following groups:
    \begin{itemize}
    \vspace{-5pt}
        \item Metric-based methods \textbf{M}:  ProtoNet~\cite{PN}, Relation Net~\cite{sung2018learning}, FSLGNN~\cite{satorras2018few}, 
    \vspace{-5pt}
        \item Optimization-based methods \textbf{O}: Meta-learner~\cite{ravi2016optimization},  MAML~\cite{finn2017model}, MetaOptNet~\cite{MetaOpt}
    \vspace{-5pt}
    \end{itemize}

    The comparison results of the metric-based methods and optimization-based methods are shown in Table~\ref{table:baseline_fsl}. One can observe that ProtoNet~\cite{PN} with PointNet~\cite{qi2017pointnet} backbone can achieve top performance at $65.31\%$ and $65.96\%$ respectively at both datasets and still has a large room for further improvement. 

    Moreover, one can conclude that metric-based methods outperform the optimization-based methods in the point cloud scenario. One possible explanation is that the optimization-based methods are more sensitive to neural network architecture and require arduous hyperparameter settings to achieve good generalization, as evidenced by recent research in 2D image domains \cite{AntoniouES19}.

    It's worth noting that ProtoNet~\cite{PN} and MetaOptNet~\cite{MetaOpt} have the same number of parameters because they employ parameter-free square Euclidean distance and SVM as meta classifier respectively. However, solving the quadratic programming of SVM in MetaOptNet~\cite{MetaOpt} is very computationally expensive. In brief, ProtoNet~\cite{PN} has a better trade-off between classification accuracy and algorithm complexity with a high inference speed.

    \begin{table*}[!thb]\centering
    \setlength{\tabcolsep}{3mm}{
    \begin{footnotesize}\scalebox{0.9}{
    \begin{tabular}{r|r|cc|cc|ccc}\thickhline
    \multicolumn{2}{c|}{\multirow{2}{*}{Method}}& \multicolumn{2}{c|}{ModelNet40-FS}& \multicolumn{2}{c|}{ShapeNet70-FS} & \multicolumn{3}{c}{5way-1shot-15query} \\\cline{3-9} 
    \multicolumn{2}{c|}{} & 5w-1s & 5w-5s & 5w-1s & 5w-5s & PN& GFLOPs & TPS    \\ \hline 
    \multirow{3}{*}{M}
    & ProtoNet~\cite{PN}& \textbf{65.31 $\pm$ 0.78} & \underline{79.04 $\pm$ 0.54} & \textbf{65.96 $\pm$ 0.81} & \textbf{78.77 $\pm$ 0.67} & \textbf{0.15M}  & \textbf{6.16}   & \textbf{118.23} \\
    & Relation Net~\cite{sung2018learning}& 64.10 $\pm$ 0.72 & 75.75 $\pm$ 0.57 & \underline{65.88 $\pm$ 0.85} & 76.25 $\pm$ 0.71 & \underline{0.28M}  & 6.50    & \underline{96.03}  \\
    & FSLGNN~\cite{satorras2018few}& 59.69 $\pm$ 0.73 & 76.06 $\pm$ 0.63 & 64.98 $\pm$ 0.84 & 76.14 $\pm$ 0.73 & 2.23M  & 10.44  & 69.02  \\ \hline
    \multirow{3}{*}{O} 
    & Meta-learner~\cite{ravi2016optimization}& 58.69 $\pm$ 0.81 & 76.60 $\pm$ 0.65 & 62.64 $\pm$ 0.91 & 73.10 $\pm$ 0.80 & 0.88M  & 6.28   & 9.14   \\
    & MAML~\cite{finn2017model}& 57.58 $\pm$ 0.89 & 77.95 $\pm$ 0.62 & 59.20 $\pm$ 0.88 & 75.10 $\pm$ 0.75 & 0.68M  & \underline{6.19}      & 38.71  \\
    & MetaOptNet~\cite{MetaOpt}& \underline{64.99 $\pm$ 0.87} & \textbf{79.54 $\pm$ 0.61} & 65.08 $\pm$ 0.89 & \underline{77.81 $\pm$ 0.75} & \textbf{0.15M}  & \textbf{6.16}   & 11.95 \\  \thickhline  
    \end{tabular}}
    \caption{\label{table:baseline_fsl} Few-shot point cloud classification results with 95\% confidence intervals on ModelNet40-FS and ShapeNet70-FS with PointNet~\cite{qi2017pointnet} as backbone. Bold denotes the best result and underline represents the second best. $M$  is the metric based methods and $O$ is  the optimization based methods. PN: Parameter Number. GFLOPs: the number of floating-point operations. TPS: inference Tasks per Second on NVIDIA 2080Ti GPU, with 15 query examples for each task.}
    \end{footnotesize}}\vspace{-13pt}
    \end{table*}
    
    \begin{table}[tb]\centering
    \setlength{\tabcolsep}{2mm}{
    \begin{footnotesize}\scalebox{0.85}{
    \begin{tabular}{c|r|cc cc} \thickhline 
    \multirow{2}{*}{Method}&\multicolumn{1}{c|}{\multirow{2}{*}{Backbone}}  & \multicolumn{2}{c}{ModelNet40-FS} & \multicolumn{2}{c}{ShapeNet70-FS} \\ 
    \cline{3-6}   &    & 5w-1s & 5w-5s  & 5w-1s  & 5w-5s    \\ \hline 
    \multirow{6}{*}{ProtoNet~\cite{PN}} & PointNet~\cite{qi2017pointnet}  & 65.31 & 79.04  & 65.96  & 78.77\\
    \small& PointNet++~\cite{qi2017pointnet++} & 64.96 &83.66      & 66.33  & 80.95\\
    \small&PointCNN~\cite{li2018pointcnn}   & 60.38 & 76.95  &64.02  &76.34   \\
    \small& RSCNN~\cite{liu2019relation}      & \underline{69.72} & \underline{84.79}  & \underline{68.66}  &\textbf{82.55}\\
    \small& DensePoint~\cite{liu2019densepoint} & 66.99 & 82.85  &65.81  & 80.74  \\ 
    \small& DGCNN~\cite{wang2019dynamic}      & \textbf{69.95} & \textbf{85.51}  & \textbf{69.03}  & \underline{82.08} \\ 
    \thickhline 
    \end{tabular}}
    \caption{\label{table:baseline_backbone} Results of ProtoNet~\cite{PN} on ModelNet40-FS and ShapeNet70-FS with different 3D backbone architectures. Bold denotes the best result and underline represents the second best.}
    \end{footnotesize}}
    \vspace{-15pt}
    \end{table}

\subsection{Influence of Backbone Architecture on FSL}
    Then we further study the influence of the backbones to point cloud FSL in ProtoNet~\cite{PN}. We select three types of current state-of-the-art 3D point-based networks including:
    \begin{itemize}
    \vspace{-5pt}
    \item Pointwise-based: PointNet~\cite{qi2017pointnet} and PointNet++~\cite{qi2017pointnet++}.
    \vspace{-5pt}
    \item Convolution-based: PointCNN~\cite{li2018pointcnn}, RSCNN~\cite{liu2019relation} and DensePoint~\cite{liu2019densepoint}.
    \vspace{-5pt}
    \item Graph-based: DGCNN~\cite{wang2019dynamic}.
    \vspace{-5pt}
    \end{itemize}
    
    For fair comparisons, we remove the last fully connected layers and train the networks from scratch with the same training strategy as suggested by these methods. After that, we feed the embedding features generated by these networks into ProtoNet~\cite{PN} for few-shot classification and the comparison results are shown in Table~\ref{table:baseline_backbone}. One can conclude that the graph-based network DGCNN~\cite{wang2019dynamic} achieves higher classification accuracy than other networks on these two datasets. The reason may be that graph-based methods dynamically update the point-wise connection graph in feature space and extracts edge features with EdgeConv layers hence can learn more discriminative features. 
    
    Therefore, we will use ProtoNet~\cite{PN} with DGCNN~\cite{wang2019dynamic} as the strong baseline for benchmarking with 3D FSL.

\vspace{-6pt}
\section{Approach}
\subsection{Overview}
\label{sec:pipeline}
    Given the baseline in the aforementioned section, 3D FSL still faces certain challenges: 1) there are strong intra-class variations and subtle inter-class differences among the support set and the query set with a small amount of data; 2) many FSL methods extract features from support and query examples independently without considering the correlations between these two sets, therefore current features learning are not discriminative enough.

\begin{figure*}[htp]
    \begin{center}
    \includegraphics[width=0.8\linewidth]{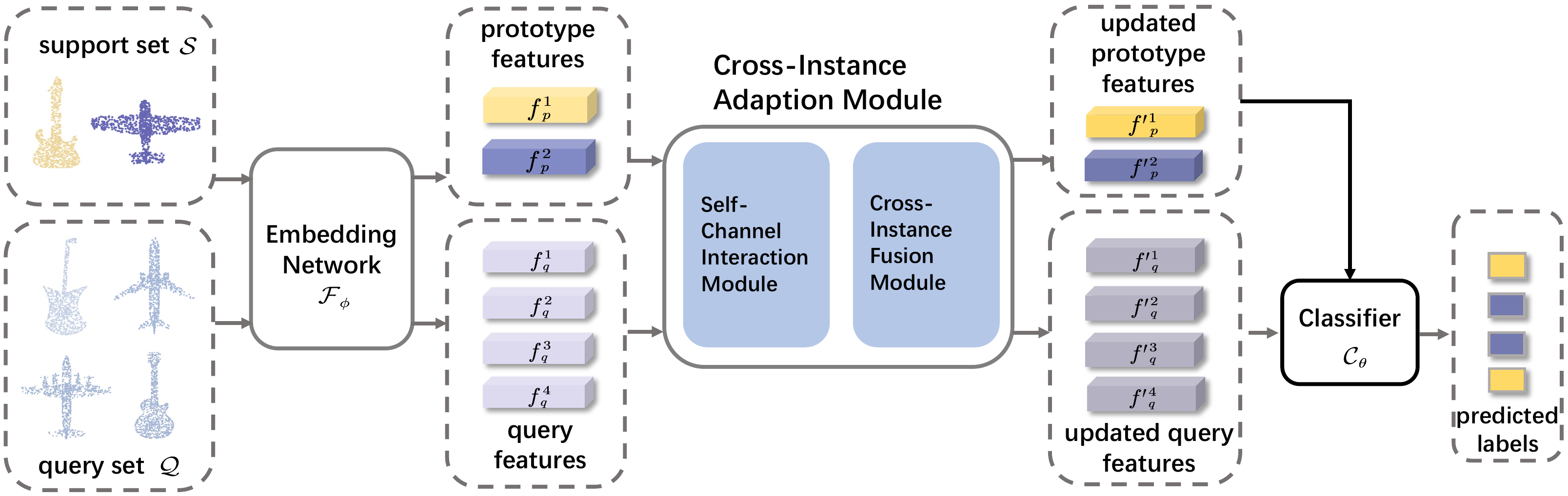}
    \vspace{-10pt}
    \end{center}
    \caption{An illustration of our proposed framework for 3D few-shot point cloud classification with an introduction in Section~\ref{sec:pipeline}.  For clarity, we only present the 2-way 1-shot 2-query setting.}
    \label{fig:pipeline}\vspace{-15pt}
    \end{figure*}

    To address these challenges, we propose a novel plug-and-play Cross Instance Adaption (CIA) module, which can be inserted into existing backbones and learning frameworks to learn more discriminative representations for the support set and query set, to be elaborated in Section \ref{subsec:cia_module}. By integrating the CIA module with the current meta-learning framework, we come up with a novel and strong network for 3D FSL classification, which is illustrated in Figure~\ref{fig:pipeline}. 

    First, the embedding module $\mathcal{F}_{\phi}$  takes support set $\mathcal{S}$ and query set $\mathcal{Q}$ as input, and maps each point cloud example $x\in \mathbb{R}^{n\times 3}$ as a feature vector $f=\mathcal{F}_{\phi}\left( x \right)$, where $f \in \mathbb{R}^{1\times d}$. Then we define the prototype feature for class $c_i$ in support set as $f_p^i=\frac{1}{|K|}\varSigma _{x_s\in c_i}\mathcal{F}_{\phi}\left( x_s \right) $ as the mean of its $K$ support examples, and the query feature for a query example $x_q^j$ as $f_q^j=\mathcal{F}_{\phi}\left( x_q^j \right)$, where $i\in[1,N]$ and $j\in[1,N_q]$.
    
\begin{figure}[htp]
    \begin{center}
       \includegraphics[width=1\linewidth]{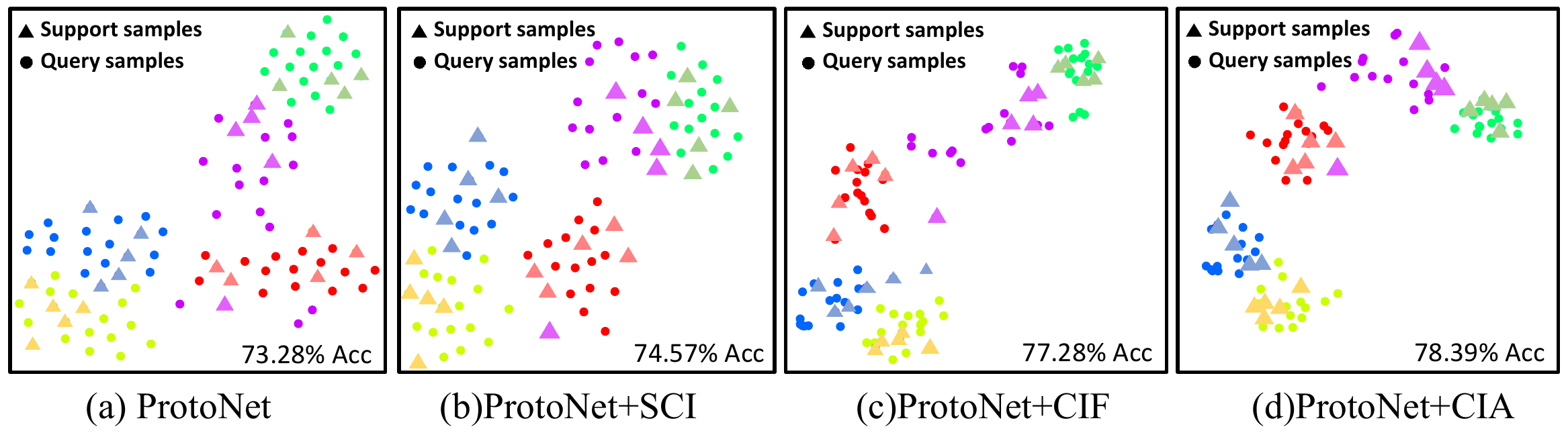}\vspace{-10pt}
    \end{center}
       \caption{The t-SNE comparison of feature distribution of support set and query set before and after using the CIA modules (SCI and CIF) in Section \ref{subsec:cia_module}. $\bigtriangleup$ stands for the support features,  $\bullet$ represents the query features.}
    \label{fig:tsne}\vspace{-15pt}
    \end{figure}

    However, the first step extracts feature from the support set and the query set separately. It also ignores the high intra-class variance and subtle inter-class differences issues in 3D FSL. Hence, the learned support and query features are not discriminative enough with a huge distribution shift, as demonstrated in Figure \ref{fig:tsne} (a). Therefore, we propose to update the prototype features ${f_p^i}$ and query feature ${f_q^j}$ by feeding them into the novel Cross Instance Adaption (CIA) module in Section \ref{subsec:cia_module}, to learn more diverse and discriminative support and query features ${f'_{p}}^i$ and ${f'_{q}}^j$, and to mitigate the distribution shift for better classification, as demonstrated in Figure \ref{fig:tsne} (d). 
    
    After that, we take Square Euclidean Distance metric function as classifier $\mathcal{C}_{\theta}$ to measure the distance between each class prototype and query examples in the feature space. The probability of predicted label $\hat{y_j}$  for  ${f'_q}^j$  as class $c_i$ is denoted as:
    \vspace{-5pt}
    \begin{equation}
    \label{eq:overview_probability}
    \vspace{-5pt}
        p\left( \hat{y_j}=c_i|{f'_q}^j \right) =\frac{\exp \left( -d\left( {f'_q}^j,{f'_p}^{c_i} \right) \right)}{\sum_{i=1}^{N}{\exp \left( -d\left( {f'_q}^j,{f'_p}^i \right) \right)}}, 
    \end{equation}
    where $d\left( .,. \right) $  is the Square Euclidean Distance, ${f'_p}^i$ and  ${f'_q}^j$ are the updated features generated by CIA module. Finally, we can get the cross-entropy loss with Equation~\ref{eq:cross_entropy}, and optimize the network end-to-end by minimizing the Equation~\ref{eq:overview_loss}:
    \vspace{-13pt}
    \begin{equation}
    \label{eq:overview_loss}
        L_{CE} = -\frac{1}{N}\frac{1}{N_q}\sum_i^{N}{\sum_j^{N_q}{\mathbbm{1} \left[ y_j=c_i \right] \log \left( p\left( \hat{y}_j=c_i|{f'_q}^j \right) \right)}},
    \end{equation}
    where $N$ and $N_q$  are the numbers of class prototypes and query examples respectively, $y_j$ is the ground truth of ${f'_q}^j$, and $\mathbbm{1}$ denotes the  Kronecker delta function.
    
\vspace{-3pt}
\subsection{Cross-Instance Adaption Module}
\label{subsec:cia_module}
    To learn more discriminative features, the Cross Instance Adaption module (CIA) consists of two modules: first, a Self-Channel Interaction (SCI) module is designed to learn diverse and discriminative features of a point clouds object by modeling channel correlations to address the issues of subtle inter-class differences, and then a Cross-Instance Fusion (CIF) module is designed to explore instance-wise interaction to address high intra-class variances issues, which can compensate the prototypical information and rectify the feature distribution by re-weighting support and query features with a meta-learner. 
    
\vspace{-10pt}
\subsubsection{Self-Channel Interaction Module}
\begin{figure}[tp]
    \begin{center}
    \includegraphics[width=1\linewidth]{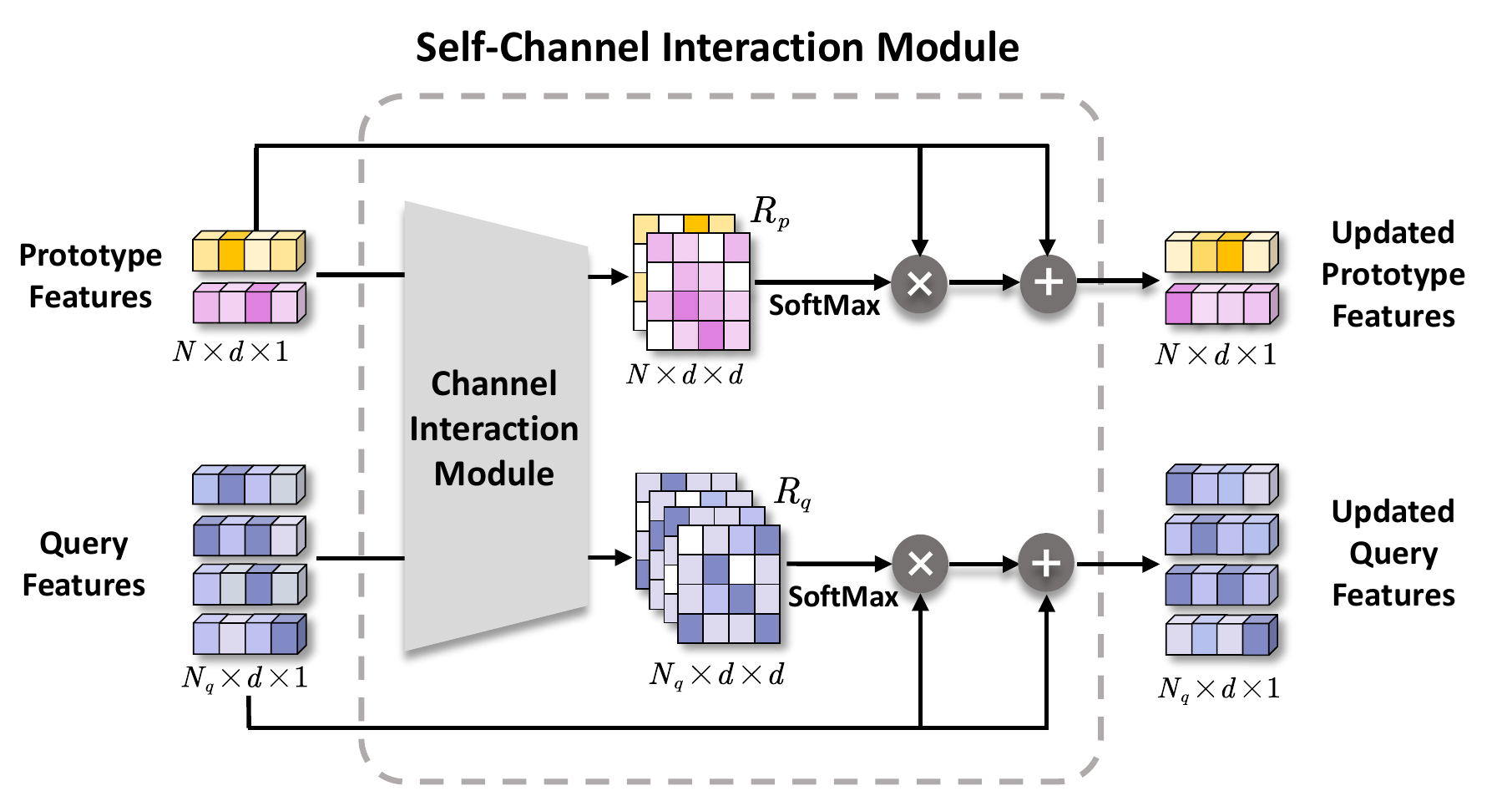}\vspace{-10pt}
    \end{center}
    \caption{An illustration of the Self-Channel Interaction Module. For clarity, we only present the 2-way 1-shot 2-query setting.}
    \label{fig:sci_module}\vspace{-15pt}
    \end{figure}

    3D FSL faces the challenge of subtle inter-class differences, for example, 'chair' and 'bench' can both have samples with handles and cushions. ~\cite{yosinski2015understanding,gao2020channel} point out that different channels can convey different semantic information. Inspired by this observation, we attempt to model channel-wise relationships to learn more diverse features to address subtle inter-class differences with an illustration in Figure~\ref{fig:sci_module}.
    
    Concretely, a query-vector $q\in \mathbb{R}^{1\times d}$ and a  key-vector $k \in \mathbb{R}^{1\times d}$ are firstly generated from the embedding feature vector $f$ with two linear embedding functions parameterized by $\varphi$ and $\gamma $ respectively. Then Channel Interaction Module is executed by using a bilinear operation between $q^T$  and $k$ to get a channel-wise relation score map:
    \vspace{-2pt}
    \begin{equation}
    \label{eq:sci_relation_score}
        R=q^Tk,~ R\in \mathbb{R}^{d\times d}.
    \end{equation}
    
    \vspace{-3pt}
    Then we normalize each column of $R$ by performing a softmax function to get the weight matrix $R'$. Specially, the $i^{th}$ value of the $j^{th}$ column in $R'$ can be calculated by:
    \vspace{-2pt}
    \begin{equation}
    \label{eq:sci_softmax}
        R_{ij}^{'}={\frac{\exp \left( -R_{ij} \right)}{\sum\limits_{k=1}^d{\exp \left( -R_{kj} \right)}}},~R'\in \mathbb{R}^{d\times d},
        \vspace{-2pt}
    \end{equation}
    where $\sum\limits_{k=1}^d{R_{kj}^{'}}=1$. After that, we use the channel-wise relation score map $R'$ to re-weight the initial feature $f$ and get vector $v$, which can be denoted by:
    \begin{equation}
    \label{eq:sci_reweight}
        v=fR', \ v \in \mathbb{R}^{1\times d}.
        \vspace{-2pt}
    \end{equation}
    
    Note that the $i^{th}$ channel of $v$, defined as $v_i=f_1R_{i1}^{'}+...+f_dR_{id}^{'}$, is the weighted sum of all channels in $f$, so higher value of $v_i$ indicates that the $i^{th}$ channel is more informative. Finally, we further combine $v$ and $f$ to compensate the discarded information and output the updated features $f'$:
    \vspace{-2pt}
    \begin{equation}
    \label{eq:sci_res}
        f'= v+f, ~f'\in \mathbb{R}^{1\times d}
    \end{equation}
    After using the SCI module, the features become more discriminative to separate different classes, as shown in Figure \ref{fig:tsne} (b). More visualization analysis of the SCI module can be found in Supplementary Material.
    
\vspace{-10pt}
\subsubsection{Cross-Instance Fusion Module}
\vspace{-3pt}
    Most existing methods~\cite{vinyals2016matching,PN,sung2018learning,oreshkin2018tadam} extract support features and query features independently, as a result, there are huge feature distribution shifts (as shown in Figure \ref{fig:tsne} (a)) exists between the support set and the query set due to the low volume of data and high intra-class variance issues. Thus, we propose a simple but effective Cross-Instance Fusion (CIF) module with a meta-network to re-weight and update support features and query features by considering their instance-wise relationships.

\begin{figure}[tp]
    \begin{center}
    \includegraphics[width=1\linewidth]{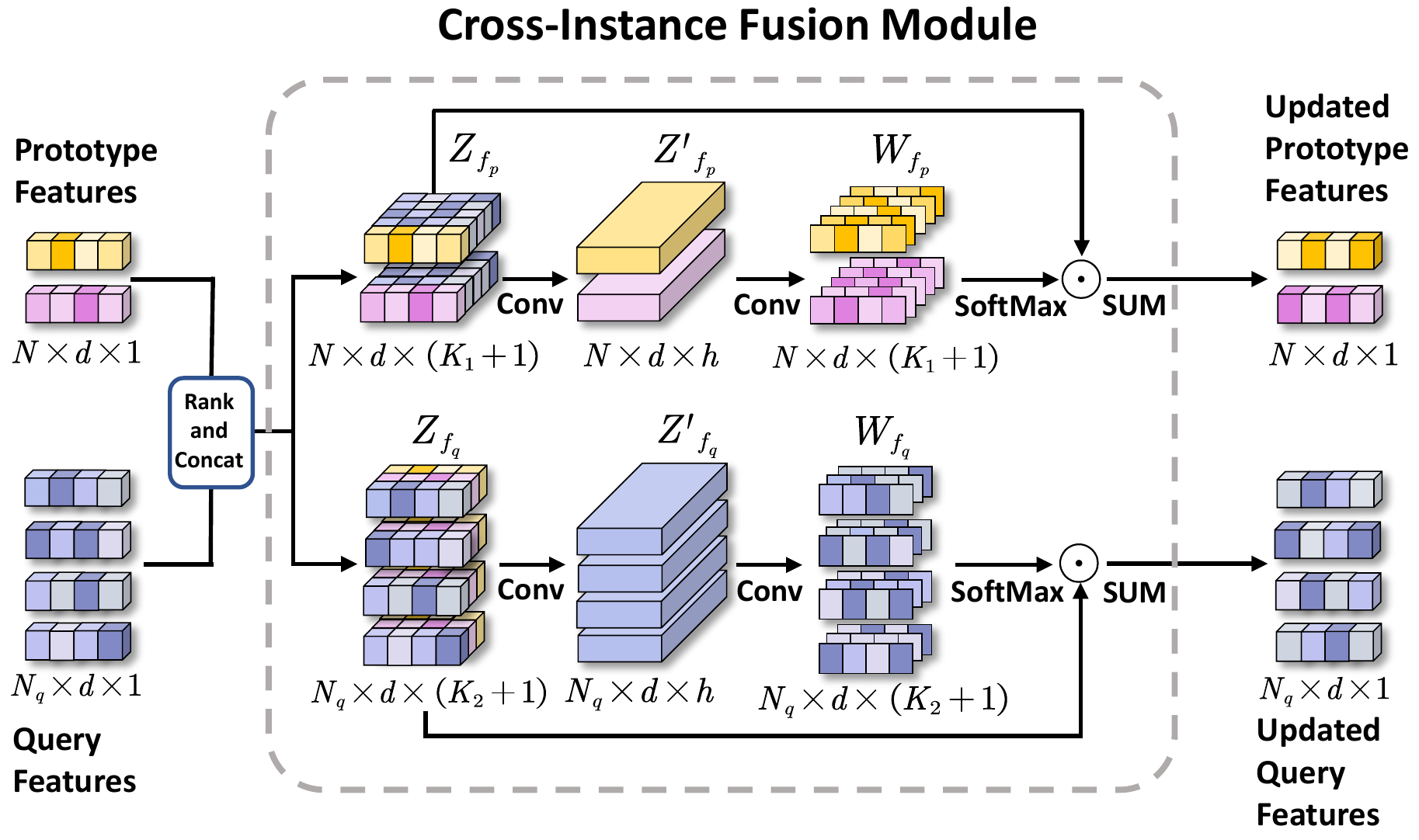}\vspace{-10pt}
    \end{center}
    \caption{An illustration of the Cross-Instance Fusion Module.  For clarity, we only present the 2-way 1-shot 2-query setting. The $\odot$ is the element-wise product. }
    \label{fig:cif_module}\vspace{-12pt}
    \end{figure}

    Specifically, as illustrated in Figure~\ref{fig:cif_module}, we first concatenate each prototype feature $f_p^i$ (where $i\in[1,N]$) to its top $K_1$ cosine similar query features, and get $Z_{f_p^i}$:
    \begin{equation}
    \label{eq:cif_proto_feature_concat}
        Z_{f_p^i}=\left[ f_p^i, f_q^{\left< top1 \right>},..., f_q^{\left< topK_1 \right>} \right], Z_{f_p^i}\in \mathbb{R}^{1\times d\times \left( K_1+1 \right)},
    \end{equation}
    where [$\cdot$] is the concatenation operation,  $d$ is the number of feature channels, $K_1 \le N_q$, and $f_q^{\left< top1 \right>}$ represents the query feature having the highest cosine similarity with prototype feature $f_p^i$. Similarly, we concatenate each query feature $f_q^j$  (where $j\in[1,N_q]$) to its top $K_2 (\le N)$ cosine similar prototype features, and get $Z_{f_q^j}$:
    \begin{equation}
    \label{eq:cif_query_feature_concat}
        Z_{f_q^j}=\left[ f_q^j, f_p^{\left< top1 \right>},..., f_p^{\left< topK_2 \right>} \right], Z_{f_q^j}\in \mathbb{R}^{1\times d\times \left( K_2+1 \right)}.
    \end{equation}
    
    We then employ two $1\times1$ convolution layers as a meta-leaner to encode the concatenated features and generate a weight matrix $W_{f}$ for $Z_{f}$. After that, we update the prototype (query) features by using the weighted sum of $Z_{f}$ instead of simple averaging, which can fuse the instance-wise information flexibly. For example, the weight matrix $W_{f_p^i}$ for $Z_{f_p^i}$ is denoted as:
    \begin{equation}
    \label{eq:cif_weight_matrix}
        W_{f_p^i}=f_2\left( f_1\left( Z_{f_p^i} \right) \right), W_{f_p^i} \in \mathbb{R}^{1\times d\times \left( K_1+1 \right)}.
    \end{equation}
    
    And the prototype feature $f_p^i$ can be updated by combining the $K_1$ concatenated features in $Z_{f_p^i}$ based on $W_{f_p^i}$:
    \begin{equation}
    \label{eq:cif_updata_feature}
         {f'_p}^i=\sum\limits\left(softmax\left(W_{f_p^i}\right) \odot Z_{f_p^i} \right), {f'_p}^i \in \mathbb{R}^{1\times d}
    \end{equation} 
    where $f_1\left( \cdot \right)$ is the first $1\times1$ Conv layer encoding the $Z_{f_p^i}$ into a $h$-dim feature interaction $Z'_{f_p^i}$, and the second $1\times1$ Conv layer $f_2\left( \cdot \right)$ is designed to adjust the dimension of interaction $Z'_{f_p^i}$ to generate the weight matrix $W_{f_p^i}$, and $\odot$ is the element-wise product.

    Similarly, we can easily get $W_{f_q^j}=f_4\left( f_3\left( Z_{f_q^j} \right) \right) \in \mathbb{R}^{1\times d\times \left( K_2+1 \right)}$ for $Z_{f_q^j}$ with two $1\times1$ Conv layers  $f_3\left( \cdot \right)$  and $f_4\left( \cdot \right)$, whose parameters are not shared with $f_1\left( \cdot \right)$  and $f_2\left( \cdot \right)$, and generate the updated query features ${f'_q}^j$.

    After using the CIF module, the distribution shift between support and query set is mitigated, as shown in Figure \ref{fig:tsne} (c). More visualization analysis of the CIF module can be found in Supplementary Material.
    
\vspace{-3pt}
\section{Experiments}
\label{sec:experiments}
    In this section, first, we introduce two datasets for few-shot 3D point cloud classification and describe the implementation details. Then, we compare the state-of-the-art FSL algorithms with DGCNN~\cite{wang2019dynamic} as backbone using CIA modules on these datasets. Finally, we conduct extensive ablation studies to evaluate the effectiveness of the CIA module when inserted into different algorithms. 
    
\vspace{-2pt}
\subsection{Dataset}
\vspace{-2pt}
\label{subsec:dataset}
    ModelNet40~\cite{ModelNet} and ShapeNetCore~\cite{ShapeNet} are two standard benchmark datasets for 3D point cloud learning. However, exiting splits of the training set and testing set are class-overlapping, which should  be disjunctive in the few-shot setting~\cite{closerlook,vinyals2016matching}. To meet the requirement, we propose new splits of ModelNet40 and ShapeNetCore, and build two benchmarks, ModelNet40-FS and ShapeNet70-FS, for 3D few-shot point cloud classification. We carefully split the dataset according to the number of instances in each category, making sure that the data distributions of the training set and testing set are similar.  Concretely, {\bf ModeNet40-FS} includes 30 training classes with 9,240 examples and 10 testing classes with 3,104 examples. {\bf ShapeNet70-FS} has a larger number of data, which totally contains 30,073 examples, 50 classes with 21,722 samples for training, and 20 classes with 8,351 samples for testing. Details of these splits are listed in Supplementary Material. 
    
    \begin{table*}[th]\centering
    \setlength{\tabcolsep}{4mm}{
    \begin{footnotesize}\scalebox{0.9}{
    \begin{tabular}{r|r|cccc} \thickhline 
    \multirow{2}{*}{Backbone} & \multicolumn{1}{c|}{\multirow{2}{*}{Method}} & \multicolumn{2}{c}{ModelNet40-FS}  & \multicolumn{2}{c}{ShapeNet70-FS} \\ \cline{3-6} 
    & \multicolumn{1}{c|}{} & 5w-1s     & 5w-5s  & 5w-1s  & 5w-5s \\ \hline
    \multirow{8}{*}{DGCNN~\cite{wang2019dynamic}} 
    & ProtoNet~\cite{PN}      & \underline{69.95 $\pm$ 0.67}     & \underline{85.51 $\pm$ 0.52} & \underline{69.03 $\pm$ 0.84}       & \underline{82.08 $\pm$ 0.72}\\
    & Relation Net~\cite{sung2018learning}      & 68.57 $\pm$ 0.73&82.01 $\pm$ 0.53 & 67.87 $\pm$ 0.86    & 77.99 $\pm$ 0.70\\
    & FSLGNN~\cite{satorras2018few}             & 61.96 $\pm$ 0.76    &80.22 $\pm$ 0.55 & 66.25 $\pm$ 0.88        & 76.20 $\pm$ 0.77  \\
    & Meta-learner~\cite{ravi2016optimization}             & 59.08 $\pm$ 0.86         &76.99 $\pm$ 0.67     & 64.53 $\pm$ 0.83  & 74.61 $\pm$ 0.72     \\
    & MAML~\cite{finn2017model}        & 62.57 $\pm$ 0.88         &77.41$\pm$ 0.73     & 64.39 $\pm$ 0.76 & 74.11 $\pm$ 0.68     \\
    & MetaOptNet~\cite{MetaOpt}  &67.05 $\pm$ 0.78 &85.05 $\pm$ 0.59     &68.27 $\pm$ 0.93  &81.06 $\pm$ 0.76     \\ \cline{2-6}
    &\multicolumn{1}{c|} {\textbf{Ours}}                   &\textbf{75.70 $\pm$ 0.74}  & \textbf{87.15 $\pm$ 0.47}  & \textbf{73.57 $\pm$ 0.81}  &\textbf{83.24 $\pm$ 0.67}  \\
    \thickhline 
    \end{tabular}}
    \caption{\label{table:baseline_benchmark} Few-shot point cloud classification results with 95\% confidence intervals on ModelNet40-FS and ShapeNet70-FS with DGCNN~\cite{wang2019dynamic} as backbone. Bold denotes the best result and underline represents the second best.}
    \end{footnotesize}}\vspace{-14pt}
    \end{table*}
    
\vspace{-2pt}
\subsection{Training and Testing Details}
\vspace{-2pt}
     We follow the standard episode-based FSL setting ~\cite{closerlook} to train and evaluate. We first train the network from scratch with 80 epochs, and each epoch contains 400 meta-training episodes and 600 validating episodes randomly sampled from the training set. Each episode consists of $N$ classes with $K$ labeled support examples and $Q$ query examples for each class, which is denoted as the $N$-way $K$-shot $Q$-query setting. Once the meta-training is ended, we test the network with 700 meta-testing episodes which are randomly sampled from the testing set with the same $N$-way $K$-shot $Q$-query setting. We average the classification results of these meta-testing episodes with 95\% confidence intervals as the final performance. Specially, we employ 5-fold cross-validation, where the training set will be randomly divided into 5 even subsets, and each subset is used once as a validation set. Finally, we report the average performance of the 5 estimations on the testing set.
     
\vspace{-2pt}
\subsection{Implementation Details}
\vspace{-1pt}
    We take DGCNN~\cite{wang2019dynamic} as the embedding network, consisting of four EdgeConv layers (64,64,128,256) and an MLP layer. More adapted details of backbone networks and FSL algorithms  can refer to Supplementary Material.  Each point cloud instance consists of 512 points sampled randomly from the CAD model surface. We use Adam optimizer with an initial learning rate of 0.0008 and gamma of 0.5. The learning rate declines every 5 epochs. We also apply random points jittering and rotating to augment data as in ~\cite{qi2017pointnet} during training. Other specific implementation details are described in the following subsections.
    \begin{table}[t]\centering
    \setlength{\tabcolsep}{3.mm}{
    \begin{footnotesize} \scalebox{0.9}{
    \begin{tabular}{r|cc cc} \thickhline
    \multicolumn{1}{c|}{\multirow{2}{*}{Method}} & \multicolumn{2}{c}{ModelNet40-FS} & \multicolumn{2}{c}{ShapeNet70-FS} \\ \cline{2-5} 
    \multicolumn{1}{c|}{}   & 5w-1s   & 5w-5s  & 5w-1s  & 5w-5s    \\ \hline 
    ProtoNet~\cite{PN}     & 69.95   & 85.51   & 69.03    & 82.08      \\
    ProtoNet~\cite{PN}+CIA      & \textbf{75.70}    & \textbf{87.15}     & \textbf{73.57}     &\textbf{83.24}   \\ \hline 
    Relation Net~\cite{sung2018learning}  & 68.57     & 82.01     & 67.87       & 77.99      \\
    Relation Net~\cite{sung2018learning}+CIA   & \textbf{70.55}   & \textbf{83.59} & \textbf{68.67}   & \textbf{78.60}   \\ \hline 
    FSLGNN~\cite{satorras2018few}    & 61.96    & 80.22     & 66.25     & 76.20       \\
    FSLGNN~\cite{sung2018learning}+CIA   & \textbf{63.81}     & \textbf{83.57} & \textbf{67.40} & \textbf{78.62}    \\ \hline 
    Meta-learner~\cite{ravi2016optimization}     & 59.08            & 76.99           & 64.53            & 74.61           \\
    Meta-learner~\cite{ravi2016optimization}+CIA & \textbf{60.55}& \textbf{77.30}  & \textbf{65.61}   & \textbf{75.01}  \\ \hline 
    MAML~\cite{finn2017model}                    & 62.57            & 77.41           & 64.39            & 74.11            \\
    MAML~\cite{finn2017model}+CIA              & \textbf{63.32}   & \textbf{78.29}  & \textbf{65.25}   & \textbf{75.03}   \\ \hline
    MetaOptNet~\cite{MetaOpt}          &67.05                & 85.05              &68.27    &81.06               \\
    MetaOptNet~\cite{MetaOpt}+CIA     &\textbf{74.70}    &\textbf{87.10}      &\textbf{72.82}       &\textbf{83.08}               \\
    \thickhline
    \end{tabular}}
	\caption{\label{table:cia_fsl}  Comparisons of the classification results after incorporating CIA Module into different FSL algorithms on ModelNet40-FS and ShapeNet70-FS with DGCNN~\cite{wang2019dynamic} as the backbone.}
    \end{footnotesize}} \vspace{-15pt}
    \end{table}

\subsection{Comparison with the Baselines}
    To verify the effectiveness of our proposed network, we first compare the classification accuracy with the aforementioned FSL baselines on two benchmark datasets, ModelNet40-FS and ShapeNet70-FS. The results in Table~\ref{table:baseline_benchmark} show that our method  exceeds other baselines by a large margin, about 5\% for 1-shot and 1.5\% for 5-shot, and outperforms other baselines on both two datasets. 

    One potential explanation is that ProtoNet is a metric-based approach that predicts labels for query examples based on the nearest square Euclidean distance with each class prototype. However, features for support and query sets are extracted independently. The CIA Module can adjust the distribution of support and query examples in the feature space by considering feature-level and  instance-level association, which could enhance the discrimination between prototypes and query examples.

\subsection{Ablative Analysis}
    In this section, we conduct ablation studies to analyze the effects of various designs of the proposed module for few-shot point cloud classification. 

    \textbf{Adding CIA Module to Different FSL Baselines. }We embed the CIA Module into metric-based and optimization-based FSL baselines to validate the generalization ability with multiple systems. All the experiments take DGCNN as the backbone for feature extraction and share the same training strategy for a fair comparison. Table~\ref{table:cia_fsl} shows the improvement of adapting the CIA Module into each algorithm. One can observe that there is an approximately 2\% consistency increase after incorporating with CIA Module, demonstrating that the proposed module can improve both metric-based and optimization-based FSL baselines significantly.
    
    \textbf{Adding CIA Module into Different Backbones. } We also verify the effectiveness of the CIA Module adapting into ProtoNet with different 3D support backbones aforementioned in Table~\ref{table:cia_backbone}. We first remove the FC layers of these 3D networks and feed the output feature vectors into the CIA Module to generate re-weighted features. Then we employ square Euclidean distance as the metric function to classify the unlabeled query samples. The comparison results in Table~\ref{table:cia_backbone} show that after including CIA Module, different state-of-the-art backbones can achieve significant performance improvement.

    \textbf{Ablation Study of CIA modules.} After that, we perform ablation studies to quantify the contribution of two proposed modules in CIA with the results listed in Table~\ref{table:cia_ablation}. We can clearly observe that both “SCI” and “CIF” could provide positive impacts and improve the performance of ProtoNet on MdoelNet40-FS and ShapeNet70-FS. Especially, the network incorporating “CIF” has about 3\% improvement for the 1-shot setting. 

    \begin{table}[t]\centering
    \setlength{\tabcolsep}{2.2mm}{
    \begin{footnotesize}\scalebox{0.9}{
    \begin{tabular}{r|c|cc cc}\thickhline
    \multirow{2}{*}{Backbone}   & \multirow{2}{*}{Method} & \multicolumn{2}{c}{ModelNet40-FS} & \multicolumn{2}{c}{ShapeNet70-FS} \\ \cline{3-6} 
    &   & 5w-1s    & 5w-5s    & 5w-1s   & 5w-5s   \\ \hline 
    \multirow{2}{*}{PointNet~\cite{qi2017pointnet}}   
    & PN    & 65.31   & 79.04   & 65.96   & 78.77      \\
    & PN+CIA   & \textbf{67.40}   & \textbf{80.31}  & \textbf{69.36}  & \textbf{80.31}   \\ \hline 
    \multirow{2}{*}{PointNet++\cite{qi2017pointnet++}} 
    & PN    & 64.96    & 83.66    & 66.33    & 80.95   \\
    & PN+CIA  &\textbf{72.42} & \textbf{84.93}  & \textbf{72.66}  & \textbf{82.65}  \\ \hline 
    \multirow{2}{*}{PointCNN~\cite{li2018pointcnn}}   
    & PN       & 60.38    & 76.95  & 64.02     &76.34   \\
    & PN+CIA     & \textbf{64.20}  &\textbf{80.86}    & \textbf{66.12}  &\textbf{ 78.06}    \\ \hline 
    \multirow{2}{*}{RSCNN~\cite{liu2019relation}}      
    & PN    & 69.72   & 84.79   & 68.66  & 82.55    \\
    & PN+CIA   & \textbf{74.23}  & \textbf{85.58}   & \textbf{73.86} & \textbf{83.85}  \\ \hline 
    \multirow{2}{*}{DensePoint~\cite{liu2019densepoint}} 
    & PN    & 66.99   & 82.85   & 65.81    & 80.74  \\
    & PN+CIA   & \textbf{72.36}   & \textbf{84.41} &\textbf{ 72.48}   & \textbf{81.84}  \\ \hline 
    \multirow{2}{*}{DGCNN~\cite{wang2019dynamic}}      
    & PN     & 69.95    & 85.51  & 69.03  & 82.08  \\
    & PN+CIA    & \textbf{75.70} & \textbf{87.15}  & \textbf{73.57}  & \textbf{83.24}   \\ 
     \thickhline
    \end{tabular}}
    \caption{\label{table:cia_backbone} Comparisons of the classification results after incorporating CIA Module into ProtoNet(PN)~\cite{PN} with different backbones on ModelNet40-FS and ShapeNet70-FS.} \vspace{-5pt}
    \end{footnotesize}}
    \end{table}   
    
    \begin{table}[t]
    \centering
    \setlength{\tabcolsep}{2.5mm}{
    \begin{footnotesize}
    \scalebox{0.9}{
    \begin{tabular}{ccc|cc cc}
    \thickhline 
    \multirow{2}{*}{ProtoNet} & \multirow{2}{*}{SCI} & \multirow{2}{*}{CIF} &  \multicolumn{2}{c}{ModelNet40-FS} & \multicolumn{2}{c}{ShapeNet70-FS} \\ \cline{4-7} 
     &   &   &  5w-1s & 5w-5s & 5w-1s & 5w-5s \\ \hline
    \checkmark &   &    &69.95       &85.51       &69.03       &82.08       \\
    \checkmark & \checkmark &      &70.51  &85.96  &69.37 &82.32  \\
    \checkmark &   & \checkmark    &73.67  & 86.69     &72.32  &82.73     \\
    \checkmark & \checkmark &\checkmark  & \textbf{75.70} &\textbf{87.15} & \textbf{73.57} &\textbf{83.24} \\      \thickhline
    \end{tabular}}
    \caption{\label{table:cia_ablation} Ablation study of CIA Module with DGCNN~\cite{wang2019dynamic} as embedding network. {"SCI" denotes the Self-Channel Interaction block and "CIF" means the  Cross-Instance Fusion block.}}
    \end{footnotesize}}  \vspace{-6pt}
    \end{table}

    \begin{table}[t] \centering
    \setlength{\tabcolsep}{3mm}{
    \begin{footnotesize} \scalebox{0.9}{
    \begin{tabular}{c|cc cc} \thickhline
    \multicolumn{1}{c|}{\multirow{2}{*}{Method}} & \multicolumn{2}{c}{ModelNet40-FS} & \multicolumn{2}{c}{ShapeNet70-FS} \\ \cline{2-5} 
    \multicolumn{1}{c|}{}   & 5w-1s   & 5w-5s  & 5w-1s  & 5w-5s    \\ \hline 
    SimpleShot~\cite{simpleshot}   & 61.87  & 78.17   & 61.58  & 74.63           \\
    SimpleShot /w SB~\cite{lssb}    & 63.33  & 76.41   & 64.45  & 73.77 \\ 
    SSL3DFSL~\cite{satorras2018few}    & 64.89  & 79.59   & 65.76  & 79.19  \\ \hline
    \textbf{Ours}   & \textbf{75.70}  & \textbf{87.15}  & \textbf{73.57}   & \textbf{83.24}  \\ 
    \thickhline
    \end{tabular}}
    \caption{\label{table:3dfsl}  The comparisons of other 3DFSL methods on Model-Net40-FS and ShapeNet70-FS with DGCNN~\cite{wang2019dynamic} as backbone.}
    \end{footnotesize}}
    \vspace{-17pt}
    \end{table}

    \textbf{Comparison with other 3DFSL methods. } 
    We further compare our proposed network with two recent 3D FSL works on ModelNet40\_FS and ShapeNet70\_FS. We follow the settings mentioned in~\cite{lssb} and~\cite{SSLFSL}, where Simpleshot is image-only low-shot recognition for 3D models' 2D RGB projections, and /w SB means shape-biased which uses auxiliary point cloud feature to learning a discriminative embedding space. For SSL3DFSL~\cite{SSLFSL}, We first train the based feature extractor with a self-supervised strategy proposed in~\cite{SSLFSL}, and then fine-tune the classifier with few samples. Table~\ref{table:3dfsl} summarizes the results on 5-way classification tasks, where our proposed baseline achieves much higher performance on both benchmarks.
    
\vspace{-3pt}
\subsection{Feature Visualization}
\vspace{-3pt}
    We use t-SNE~\cite{maaten2008visualizing} to visualize the feature distribution at a 5way-5shot-15query setting on ModelNet40\_FS with DGCNN~\cite{wang2019dynamic} as the backbone. Figure~\ref{fig:tsne} (a) corresponds to the features of ProtoNet without the CIA module achieving an accuracy of 73.28\%, (b) and (c) are the results of incorporating SCI and CIF, achieving 74.57\% and 77.28\% respectively, while (d) has better performance of 78.39\% after equipped with CIA module.  Note that the learning support and query features of ProtoNet~\cite{PN} are dispersed with a huge distribution shift. After using the SCI module, the support features tend to move to the centrality to better differentiate from other classes, as shown in (b) and (d). Moreover, the distribution shift between support and query sets is mitigated, as shown in (c) and (d). 

\vspace{-5pt}
\section{Conclusion}
 \vspace{-3pt}
    In this paper, we study 3D few-shot learning (FSL) in a systematic manner for the first time. We first empirically study and analyze the state-of-the-art 2D FSL algorithms in the context of point cloud classification and explore the influence of different backbone architectures. We then propose a strong baseline for benchmarking 3D FSL using ProtoNet with DGCNN. Furthermore, to alleviate the high intra-class variance and subtle inter-class differences between the support set and the query set, we also propose a plug-and-play Cross Instance Adaption (CIA) module consisting of a Self-Channel Interaction (SCI) module and a Cross-Instance Fusion (CIF) module, which can generate more discriminative features. For objective evaluation, we construct two benchmarks ModelNet40-FS and ShapeNet70-FS for 3D point cloud classification. Finally, extensive experiments show that the proposed CIA module has a significant performance improvement with different FSL algorithms and could be adapted well into different backbones.
    
    \textbf{Acknowledgement:} This work is supported by the National Natural Science Foundation of China (No. 62071127 and U1909207), Shanghai Pujiang Program (No.19PJ1402000), Shanghai Municipal Science and Technology Major Project (No.2021SHZDZX0103), Shanghai Engineering Research Center of AI Robotics and Engineering Research Center of AI Robotics, Ministry of Education in China, and Agency for Science, Technology, and Research (A*STAR) under its AME Programmatic Funding Scheme (Project A18A2b0046)

{

\small
\bibliographystyle{ieee_fullname}
\bibliography{egpaper}

\newpage


\section*{Supplementary material}

# Supplementary Material for
# What Makes for Effective Few-shot Point Cloud Classification?

Chuangguan Ye[1], Hongyuan Zhu[2], Yongbin Liao[1], Yanggang Zhang[1], Tao Chen[1], Jiayuan Fan[3]
[1]School of Information Science and Technology, Fudan University, China
[2]Institute for Infocomm Research, A*STAR, Singapore
[3]Academy for Engineering and Technology, Fudan University, China
{cgye19,19210720121,ygzhang19,eetchen,jyfan}@fudan.edu.cn, hongyuanzhu.cn@gmail.com

## A. Details about 3D Backbones

To study the influence of different backbone architectures on FSL, we select three types of current state-of-the-art 3D networks as the support backbones for features extraction. In this section, we will introduce more structural details about backbones employed in Section 3 and 5. The input point cloud instances consist of 512 points with 3d coordinates and the backbones output a feature vector with 1024 dimensions.

**Pointwise MLP Networks:** **PointNet** contains five MLP layers (64,64,64,128,1024) with learnable parameters, and batch normalization is used for all MLP layers with ReLU. After that, we use maxpooling function to aggregate a global feature vector. Note that we remove the transform layers in original PointNet [10] framework for simplicity and efficiencies. **PointNet++** consists of 3-level PointNet Set Abstractions with single scale grouping (SSG), which have the same stetting in [11]. We remove the fully connected (FC) layers and take the last PointNet Set Abstraction's output as the global feature vector.

**Convolution Networks:** There are 4 X-conv layers (48,96,192,384) with ReLU in **PointCNN**. The last X-conv layer outputs a 384-dimension feature vector and we use 2 FC layers (512,1024) to extend the dimension to 1024 for fair comparisons. **RSCNN** contains 3 RC-Conv layers (128,512,1024) with single scale neighborhood (SSN) grouping. Other settings of RC-Conv are same with [6]. **DensePoint** contains 3 P-Conv layers, 2 P-Pooling layers and 1 global pooling layer. The settings of these layers are same with [5]. We remove the FC layers in RSCNN and DensePoint for outputting the 1024-dimension global feature vectors too.

**Graph-based Network:** **DGCNN** is the embedding network of our baseline for 3DFSL, consisting of 4 Edge-Conv layers (64,64,128,256). The outputs of each Edge-Conv will be concatenated as a 512-dimension feature map. Then the feature map will be fed into an MLP layer to extend its dimension to 1024. At last, a maxpooling function is used to aggregate the global features and outputs a 1024-dimension feature vector. Figure 6 illustrates the network architecture of DGCNN.

## B. Details about FSL Baselines

In this section, we will introduce more adapting details about FSL algorithms for few-shot point cloud classification in Section 3 and 5.

**Metric-based methods:** We take Squared Euclidean Distance as metric function and use Cross-Entropy loss in ProtoNet. For RelationNet, we first construct support-query pair features $f_{sq} \in \mathbb{R}^{2 \times 1024}$ by concatenating the support feature vector $f_s \in \mathbb{R}^{1 \times 1024}$ and query feature vector $f_q \in \mathbb{R}^{1 \times 1024}$. Then the pair features $f_{sq}$ are fed into a relation module, which contains two convolutional blocks ((1x1 conv 128 filters, BN, ReLU), (1x1 conv 1 filters, BN, ReLU)) and two FC layers (128,L) (L denotes the number of classes at a meta-task). After that, relation module outputs the predicted relation score and Mean Square Error loss is used to regress relation score to ground truth. For FSLGNN, we construct 4 GNN layers with the same setting in [15], and use Cross-Entropy loss as loss function.

**Optimization-based methods:** We use two FC layers (256, K) as the classifier in Meta-learner and MAML. Batch normalization is used for the first FC layer with ReLU and Dropout operation. There is a 2-layer LSTM in Meta-learner following the same configuration in [12], where the first layer is a normal LSTM and the second layer is meta-learner LSTM for gradient state updating. The meta learning rate in MAML is 0.1 for 5-shot and 0.5 for 1-shot. The classification head in MetaOptNet is the multi-class SVM presented in [1]. We also take Cross-Entropy loss as loss function in these methods.

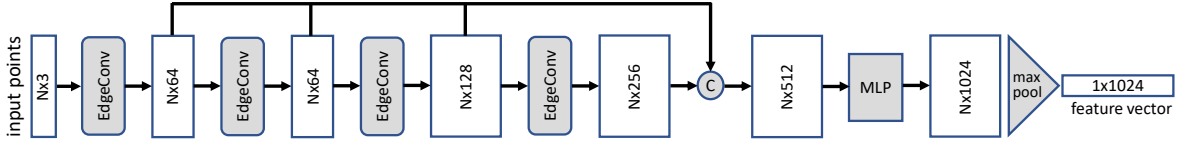

Figure 6. The architecture of DGCNN for feature embedding. The details of EdgeConv could be find in [19].

## C. Details about Channel Interaction Module

The architecture of Channel Interaction Module is shown in Figure 7. Query-vector $q$ and key-vector $k$ are generated from the input feature $f$ with two linear embedding functions $\varphi$ and $\gamma$. Then the channel relation score map can be denoted as $R = q^T k$. After that, we can obtain the reweighted feature $v = fR'$, where $R' = soft\max(R)$. Finally, for compensating the discarded information, we combine $v$ and $f$ to get updated feature $f' = v + f$.

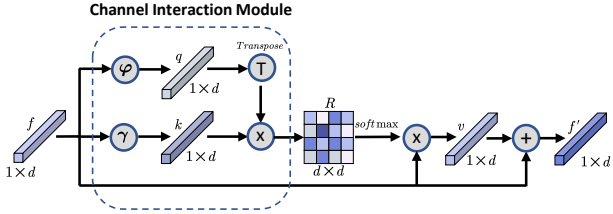

Figure 7. Details of Channel Interaction Module in Self-Channel Interaction Module introduced in Section 4.2.1.

## D. Details about Cross Instance Fusion Module

Cross-Instance Fusion (CIF) module is proposed to address the law-data and high intra-class variances issues in Section 4.2.2, which can enrich prototypical information and rectify feature distribution by fusing prototype features and query features with a meta-learner.

As illustrated in Figure 8, for updating prototype features $f_p$, we first concatenate each prototype feature with $K_1$ query features with the highest cosine similarity and get the concatenated feature $Z_{f_p}$. Then we employ two simple 1×1 Conv layers as a meta-learner to learn cross-instance interactions and output updated prototype features $f'_p$. Concretely, the first layer is designed to encode $Z_{f_p}$ to generate a $d$-dim feature interaction $Z'_{f_p}$, and the second layer is used to adjust the interaction's dimension so as to generate a reweight matrix $W_{f_p}$ for $Z_{f_p}$. Finally, we update the prototype features by fusing the concatenated feature $Z_{f_p}$ based on $W_{f_p}$. Similarly, we could also easily update the query features $f_q$ by CIF module. Furthermore, we use the validation set to determine the value of $K_1$ and $h$, and set them to 45 and 64 respectively according to the results shown in Figure 9.

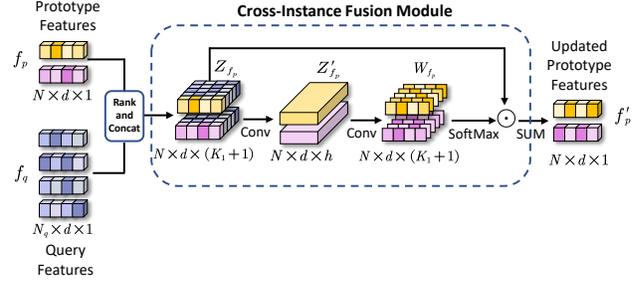

Figure 8. The illustration of updating prototype features by CIF module. Here we set $K_1 = N_q$. $\odot$ is the instance-wise product.

## E. Extra Experimental Results

We also conduct extra experiments to further explore the effects of our proposed network in two scenarios, including fine-grained few-shot classification and 2D image few-shot classification.

**Fine-Grained Few-Shot Classification.** We study the proposed network in a fine-grained classification scenario to evaluate its ability to distinguish similar categories. We first train the baselines and the proposed network on meta-train data of ShapeNet70-FS, and test them on seven subcategories of "Airplane" in meta-test data under a 5way-1shot-15query setting. The mean accuracy of each class are listed in Table 9. One can see that, our proposed network outperforms other baselines most of the time and improve the mean accuracy more than 3%.

**Ablation Studies of Residual Design in SCI Module.** We design the SCI as a residual update to compensate the discarded information, because the Softmax operation can highlight the weight of discriminative channels, but it also may discard some information from the original features. The results of ablation studies listed in Table 10 show that this residual design can gain performance improvement.

**Comparison Results of More-way k-shot Setting.** We conduct the experiments on ShapeNet70-FS with N={5,10,15}, K={1,5}, and the results are in Table 11. A larger N-way setting is more challenging with significantly performance dropping. One possible explanation is that larger classes with few support examples make training harder, leading to obscurer class boundaries.

**Effects of CIA for 2D FSL.** While the CIA model is designed for few-shot point cloud classification task, we also

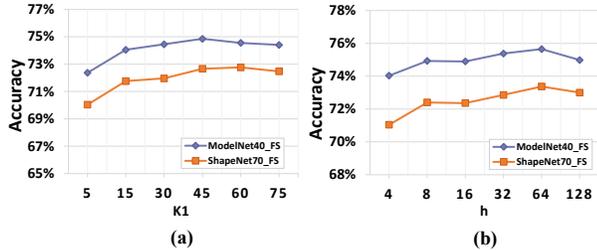

Figure 9. (a) and (b) are the ablative results of different values of $K_1$ and $h$ in Cross-Instance Fusion module respectively. We conduct these ablation experiments under the 5way-1shot-15query scenario on ModelNet40_FS and ShapeNet70_FS.

| Methods | Airline | Jet | Fighter | Swept Wing | Propeller Plane | Bomber | Delta Wing | Mean |
|---|---|---|---|---|---|---|---|---|
| ProtoNet [16] | 49.59 | 25.21 | 27.64 | 41.30 | 35.76 | 20.73 | 46.24 | 35.22 |
| RelationNet [17] | 49.28 | 24.55 | 30.36 | 41.03 | 28.26 | 24.05 | 44.79 | 34.63 |
| FSLGNN [15] | 48.53 | 23.56 | 21.01 | 31.75 | 24.29 | 23.58 | **58.49** | 33.08 |
| Meta-learner [12] | 29.76 | 21.92 | 30.94 | 27.83 | 27.02 | 24.96 | 30.22 | 27.53 |
| MAML [2] | **56.03** | **26.65** | 34.37 | 21.90 | 16.43 | 25.30 | 15.71 | 28.12 |
| MetaOptNet [4] | 46.81 | 24.72 | 29.05 | 39.74 | 24.42 | 25.17 | 38.23 | 32.60 |
| **Ours** | 48.23 | 25.51 | **35.16** | **41.88** | **46.63** | **26.06** | 48.02 | **38.80** |

Table 9. 5way-1shot-15query classification results (accuracy %) on fine-grained classes in ShapeNet70-FS.

|  | ModelNet40-FS | | ShapeNet70-FS | |
|---|---|---|---|---|
|  | 5w-1s | 5w-5s | 5w-1s | 5w-5s |
| w/o Res | 74.64 | 86.81 | 72.98 | 82.87 |
| w/ Res | **75.70** | **87.15** | **73.57** | **83.24** |

Table 10. The ablation study of residual design in SCI module.

|  | ShapeNet70-FS | | | | | |
|---|---|---|---|---|---|---|
|  | 5w-1s | 5w-5s | 10w-1s | 10w-5s | 15w-1s | 15w-5s |
| ProtoNet | 65.96 | 78.77 | 50.57 | 67.29 | 43.15 | 59.39 |
| RelationNet | 65.88 | 76.25 | 50.93 | 63.14 | 43.04 | 53.31 |
| MetaOpt | 65.08 | 77.81 | 48.97 | 64.50 | 40.83 | 56.10 |
| Ours | **69.36** | **80.31** | **54.26** | **67.69** | **47.38** | **60.48** |

Table 11. The comparison results under larger-way settings.

study the effects of CIA model for the case of 2D image few-shot classification on *mini*Imagenet and *tiered*Imagenet with ResNet12 as backbone. The results shown in Tabel 12 indicate that the CAI model also can improve the classification performance of ProtoNet [16], especially for 1-shot setting, and achieves competitive performance compared with state-of-the-art 2D FSL approaches.

## F. More Visualization Analysis

**Visualization Analysis of Feature Heatmap.** We further visualize the feature heatmap of point cloud instances to qualitatively evaluate the proposed modules in Section 4.2. Figure 10 are the comparative results before and after incorporating SCI module and CIF module respectively. Deeper color means higher feature responding in this region. We could observe that the SCI module pays more attention on the discriminative fine-grained parts of differ-

| Backbone | Method | *mini*Imagenet | | *tiered*Imagenet | |
|---|---|---|---|---|---|
|  |  | 5w-1s | 5w-5s | 5w-1s | 5w-5s |
| ResNet12 | SNAIL [8] | 55.71 | 68.88 | - | - |
|  | TADAM [9] | 58.50 | 76.70 | - | - |
|  | ECM [13] | 59.00 | 77.46 | 63.99 | 81.97 |
|  | TPN [7] | 59.46 | 75.65 | 59.91 | 73.30 |
|  | MetaOptNet [4] | 62.64 | 78.63 | 65.99 | 81.56 |
|  | CAN [3] | **63.85** | 79.44 | 69.89 | **84.23** |
|  | ProtoNet [16] | 60.37 | 79.02 | 65.65 | 83.40 |
|  | ProtoNet [16]+CIA | 63.05 | **80.02** | **70.10** | 83.73 |

Table 12. Comparisons of the classification results after incorporating CIA Module into ProtoNet [16] on *mini*Imagenet [18] and *tiered*Imagenet [14] with ResNet12 as backbone.

ent classes,such the "cap" of Bottle and the "legs" of Stool, while the CIF module could activate more diverse regions, which could help to generate more informative features. More comparative visualizations are shown in Figure 11 and 12. One can see that, the CIA module can highlight more discriminative parts and structures, which enriches the information learned from point cloud instances.

## G. Dataset Split

**ModelNet40-FS** is a new split of ModelNet40, containing 30 training classes with 9,204 examples and 10 disjoint testing classes with 3,104 examples. Statistics of ModelNet40-FS are reported in Table 13, and details of training set split and testing set split are listed in Table 15.

|  | Train | Test | Total |
|---|---|---|---|
| Classes | 30 | 10 | 40 |
| Instances | 9,204 | 3,104 | 12,308 |

Table 13. Statistics of ModelNet40-FS dataset.

**ShapeNet70-FS** is adapted from ShapeNetCore, including 50 base classes from 34 categories and 20 novel classes from 14 categories. Statistics of the ShapeNet70-FS are reported in Table 14, and details of training set split and testing set split are listed in Table 16 and Table 17 respectively.

|  | Train | Test | Total |
|---|---|---|---|
| Categories | 34 | 14 | 48 |
| Classes | 50 | 20 | 70 |
| Instances | 21,722 | 8,351 | 30,073 |

Table 14. Statistics of ShapeNet70-FS dataset.

| Training Set | | | | | | Testing Set | |
|---|---|---|---|---|---|---|---|
| Class | Num | Class | Num | Class | Num | Class | Num |
| chair | 989 | car | 297 | radio | 124 | bookshelf | 672 |
| sofa | 780 | desk | 286 | xbox | 123 | vase | 575 |
| airplane | 725 | dresser | 286 | bathtub | 156 | bottle | 435 |
| bed | 615 | glass_box | 271 | lamp | 144 | piano | 331 |
| monitor | 565 | guitar | 255 | stairs | 144 | night_stand | 286 |
| table | 492 | bench | 193 | door | 129 | range_hood | 215 |
| toilet | 444 | cone | 187 | stool | 110 | flower_pot | 169 |
| mantel | 384 | tent | 183 | wardrobe | 107 | keyboard | 165 |
| tv_stand | 367 | laptop | 169 | cup | 99 | sink | 148 |
| plant | 339 | curtain | 157 | bowl | 84 | person | 108 |

Table 15. The training and testing classes of ModelNet40-FS dataset.

| Training Set | | | | | |
|---|---|---|---|---|---|
| ID | Class | Num | ID | Class | Num |
| 04256520 | sofa | 1520 | 04037443 | race car | 323 |
| 03179701 | desk | 1226 | 20000011 | garage cabinet | 307 |
| 04401088 | phone | 1089 | 03948459 | handgun | 307 |
| 02738535 | armchair | 1051 | 04285965 | sport utility | 300 |
| 02924116 | bus | 939 | 03928116 | piano | 239 |
| 02808440 | bathtub | 856 | 02818832 | bed | 233 |
| 02992529 | radiotelephone | 831 | 04330267 | stove | 218 |
| 03891251 | park bench | 823 | 03100240 | convertible | 208 |
| 03063968 | coffee table | 763 | 04285008 | sports car | 197 |
| 20000027 | club chair | 748 | 02880940 | bowl | 186 |
| 02858304 | boat | 741 | 03141065 | cruiser | 181 |
| 04250224 | sniper rifle | 717 | 02961451 | carbine | 172 |
| 03046257 | clock | 651 | 04004475 | printer | 166 |
| 03991062 | pot | 602 | 03761084 | microwave | 152 |
| 03593526 | jar | 596 | 04225987 | skateboard | 152 |
| 03237340 | dresser | 482 | 04460130 | tower | 133 |
| 04380533 | table lamp | 464 | 20000020 | cantilever chair | 125 |
| 03642806 | laptop | 460 | 02801938 | basket | 113 |
| 04166281 | sedan | 429 | 02814533 | beach wagon | 108 |
| 03624134 | knife | 424 | 02946921 | can | 108 |
| 20000037 | rectangular table | 421 | 03938244 | pillow | 96 |
| 03119396 | coupe | 418 | 03594945 | jeep | 95 |
| 04373704 | swivel chair | 398 | 03207941 | dishwasher | 93 |
| 20000010 | desk cabinet | 356 | 04099429 | rocket | 85 |
| 03790512 | motorcycle | 337 | 02773838 | bag | 83 |

Table 16. The training classes of ShapeNet70-FS dataset. "ID" corresponds to WordNet synset offset.

| Testing Set | | | | | |
|---|---|---|---|---|---|
| ID | Class | Num | ID | Class | Num |
| 03211117 | display | 1093 | 03337140 | file cabinet | 298 |
| 02690373 | airline | 1054 | 20000001 | swept wing | 271 |
| 03467517 | guitar | 797 | 03797390 | mug | 214 |
| 03325088 | faucet | 744 | 04554684 | washer | 169 |
| 03595860 | jet | 675 | 03513137 | helmet | 162 |
| 03335030 | fighter | 597 | 04012084 | propeller plane | 137 |
| 02876657 | bottle | 498 | 02867715 | bomber | 130 |
| 02871439 | bookshelf | 452 | 03174079 | delta wing | 121 |
| 04468005 | train | 389 | 02942699 | camera | 113 |
| 02747177 | ashcan | 343 | 03710193 | mailbox | 94 |

Table 17. The testing classes of ShapeNet70-FS dataset. "ID" corresponds to WordNet synset offset.

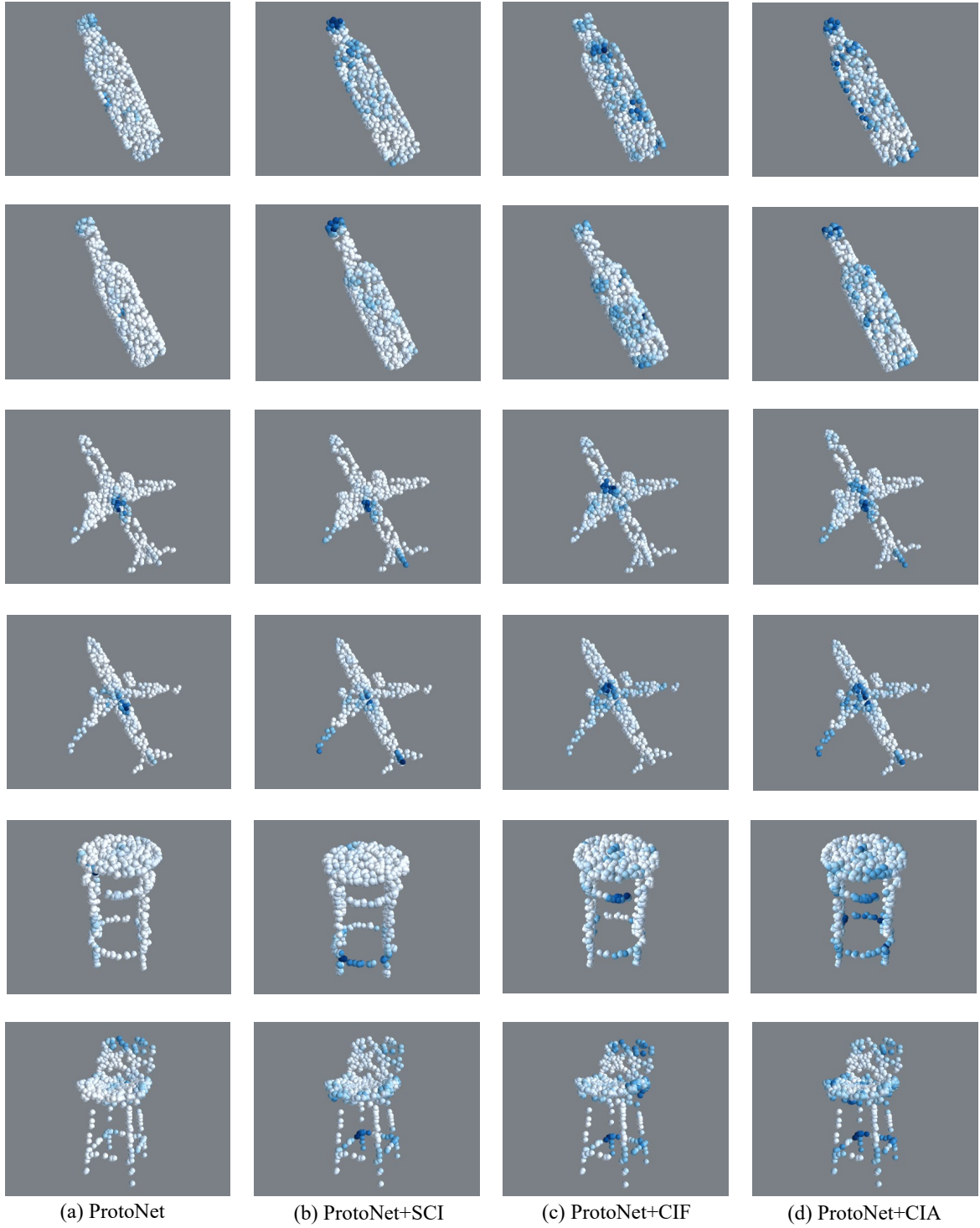

Figure 10. The heatmap of point cloud instances before and after using CIA module (SCI and CIF). Deeper color means higher feature responding in this region. The classes of each row are 'Bottle','Airplane' and 'Stool' respectively.

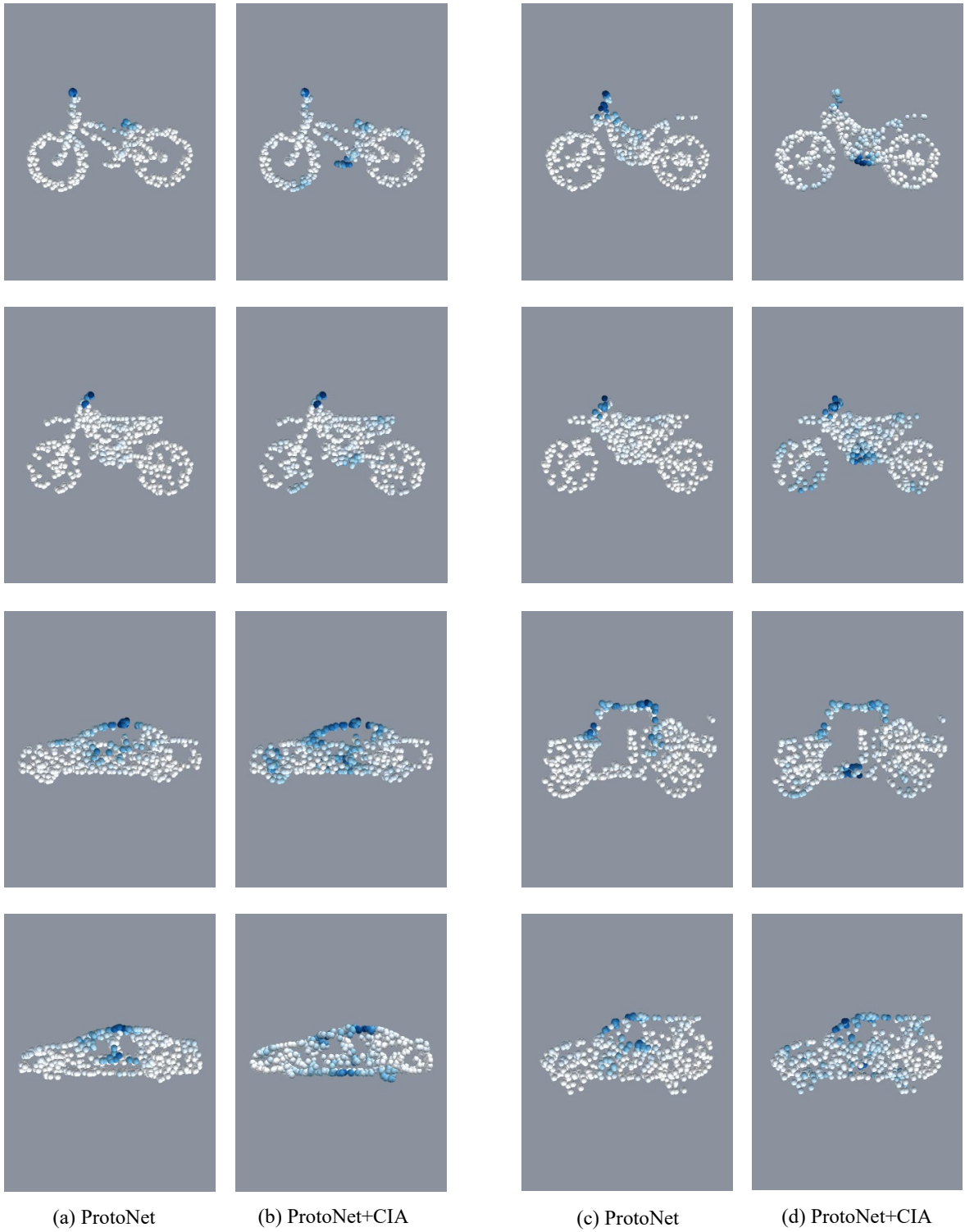

| (a) ProtoNet | (b) ProtoNet+CIA | (c) ProtoNet | (d) ProtoNet+CIA |

Figure 11. The heatmap of point cloud instances. Deeper color means higher feature responding in this region. Column(a) and column(c) are the results of ProtoNet. Column(b) and column(d) are the results of ProtoNet incorporating with CIA module.

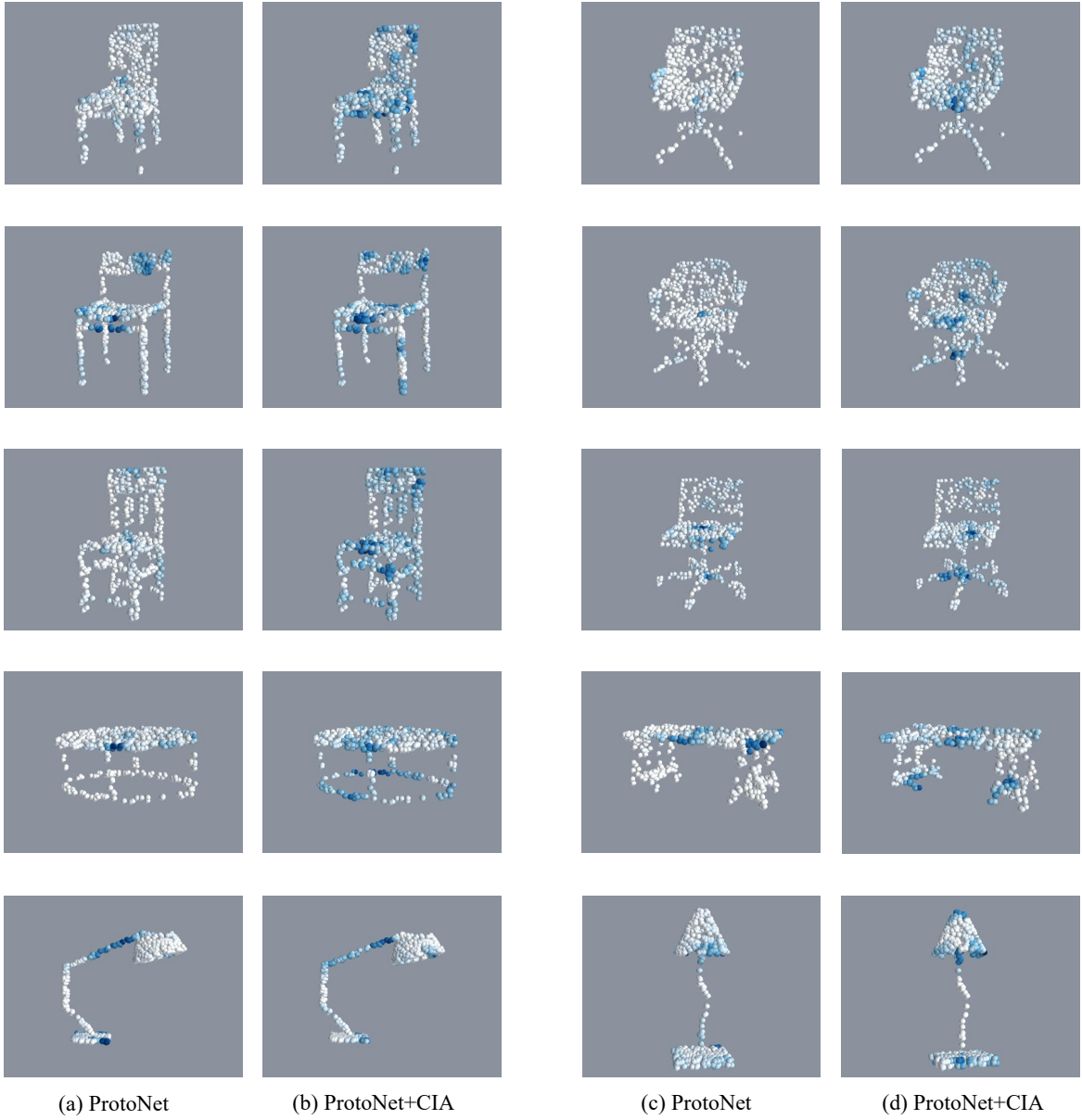

Figure 12. The heatmap of point cloud instances. Deeper color means higher feature responding in this region. Column(a) and column(c) are the results of ProtoNet. Column(b) and column(d) are the results of ProtoNet incorporating with CIA module.



}

\end{document}